\PassOptionsToPackage{unicode}{hyperref}
\PassOptionsToPackage{hyphens}{url}
\documentclass[
  11pt,
]{article}
\usepackage{xcolor}
\usepackage[margin=1in]{geometry}
\usepackage{amsmath,amssymb}
\usepackage{iftex}
\ifPDFTeX
  \usepackage[T1]{fontenc}
  \usepackage[utf8]{inputenc}
  \usepackage{textcomp} 
\else 
  \usepackage{unicode-math} 
  \defaultfontfeatures{Scale=MatchLowercase}
  \defaultfontfeatures[\rmfamily]{Ligatures=TeX,Scale=1}
\fi
\usepackage{lmodern}
\ifPDFTeX\else
\fi
\IfFileExists{upquote.sty}{\usepackage{upquote}}{}
\IfFileExists{microtype.sty}{
  \usepackage[]{microtype}
  \UseMicrotypeSet[protrusion]{basicmath} 
}{}
\usepackage{setspace}
\makeatletter
\@ifundefined{KOMAClassName}{
  \IfFileExists{parskip.sty}{%
    \usepackage{parskip}
  }{
    \setlength{\parindent}{0pt}
    \setlength{\parskip}{6pt plus 2pt minus 1pt}}
}{
  \KOMAoptions{parskip=half}}
\makeatother
\usepackage{longtable,booktabs,array}
\usepackage{calc} 
\usepackage{etoolbox}
\makeatletter
\patchcmd\longtable{\par}{\if@noskipsec\mbox{}\fi\par}{}{}
\makeatother
\IfFileExists{footnotehyper.sty}{\usepackage{footnotehyper}}{\usepackage{footnote}}
\makesavenoteenv{longtable}
\usepackage{graphicx}
\makeatletter
\newsavebox\pandoc@box
\newcommand*\pandocbounded[1]{
  \sbox\pandoc@box{#1}%
  \Gscale@div\@tempa{\textheight}{\dimexpr\ht\pandoc@box+\dp\pandoc@box\relax}%
  \Gscale@div\@tempb{\linewidth}{\wd\pandoc@box}%
  \ifdim\@tempb\p@<\@tempa\p@\let\@tempa\@tempb\fi
  \ifdim\@tempa\p@<\p@\scalebox{\@tempa}{\usebox\pandoc@box}%
  \else\usebox{\pandoc@box}%
  \fi%
}
\def\fps@figure{htbp}
\makeatother
\NewDocumentCommand\citeproctext{}{}

\makeatletter
 \let\@cite@ofmt\@firstofone
 \def\@biblabel#1{}
 \def\@cite#1#2{{#1\if@tempswa , #2\fi}}
\makeatother
\newlength{\cslhangindent}
\newlength{\csllabelwidth}
\newenvironment{CSLReferences}[2] 
 {\begin{list}{}{%
  \setlength{\itemindent}{0pt}
  \setlength{\leftmargin}{0pt}
  \setlength{\parsep}{0pt}
  \ifodd #1
   \setlength{\leftmargin}{\cslhangindent}
   \setlength{\itemindent}{-1\cslhangindent}
  \fi
  \setlength{\itemsep}{#2\baselineskip}}}
 {\end{list}}
\usepackage{calc}

\newcommand{\CSLLeftMargin}[1]{\parbox[t]{\csllabelwidth}{\strut#1\strut}}
\newcommand{\CSLRightInline}[1]{\parbox[t]{\linewidth - \csllabelwidth}{\strut#1\strut}}

\providecommand{\tightlist}{%
  \setlength{\itemsep}{0pt}\setlength{\parskip}{0pt}}
\usepackage{booktabs}
\usepackage{longtable}
\usepackage{array}
\usepackage{caption}
\usepackage{hyperref}
\hypersetup{colorlinks=true,linkcolor=blue,citecolor=blue,urlcolor=blue}
\usepackage{bookmark}
\IfFileExists{xurl.sty}{\usepackage{xurl}}{} 
\hypersetup{
  pdftitle={Fine-tuning Whisper for Pashto ASR: strategies and scale},
  pdfauthor={Hanif Rahman},
  hidelinks,
  pdfcreator={LaTeX via pandoc}}

\title{Fine-tuning Whisper for Pashto ASR: strategies and scale}
\author{Hanif Rahman\footnote{Independent Researcher.
  hanif@hanifrahman.com}}
\date{2026}

\begin{document}
\maketitle
\begin{abstract}
Pashto is primarily an oral language: poetry, proverbs, and cultural
knowledge are transmitted through recitation and conversation rather
than text. Despite being one of CommonVoice's largest language
collections, Pashto is absent from Whisper's 680,000-hour pre-training
corpus, leaving off-the-shelf models unusable for Pashto transcription.
Off-the-shelf Whisper achieves word error rates above 100\% on
CommonVoice Pashto without fine-tuning, because all model sizes output
Arabic, Dari, or Urdu script rather than Pashto text {[}1{]}. This paper
compares four fine-tuning strategies for whisper-base on CommonVoice
Pashto v20: vanilla full fine-tuning, LoRA (rank 64), frozen-encoder
(2/6 layers), and multistage Urdu→Pashto transfer. It then extends
vanilla fine-tuning to whisper-small and whisper-large-v3-turbo on the
larger CommonVoice Pashto v24 corpus (113 hours). Vanilla fine-tuning
achieves WER 21.22\% on CV20, 33.36 pp better than LoRA r=64, 14.76 pp
better than frozen-encoder fine-tuning, and 44.56 pp better than Urdu
transfer. Frozen-encoder fine-tuning degrades performance because
whisper-base has only 6 encoder layers: at this depth the layer-function
separation that motivates freezing does not hold, and freezing removes a
third of trainable capacity while preventing the lower layers from
adapting to Pashto's retroflex and pharyngeal phonemes. Urdu→Pashto
transfer fails due to an unverified intermediate checkpoint,
phonological mismatch, and insufficient training at stage 2. On CV24,
whisper-small achieves WER 24.89\%, 2.24 pp better than whisper-base at
3.3× the parameter count; whisper-large-v3-turbo achieves WER 23.37\%, a
further 1.52 pp at 3.3× more parameters. The diminishing returns at this
data scale indicate whisper-small is the practical optimum for 113-hour
Pashto training. Online data augmentation (speed perturbation and noise
injection) provides a 7.25 pp WER benefit over matched training without
augmentation. Error analysis on 100-sample subsets reveals that the
dominant failure modes are word-final morphological suffix confusion
(masculine ی vs.~feminine ه) and retroflex-phoneme substitutions
involving the Pashto-unique consonant څ /ts/, which are absent from
Arabic, Persian, and Urdu and therefore poorly represented in Whisper's
pre-training data. Fine-tuned checkpoints, the pre-augmented training
corpus (\texttt{ihanif/pashto\_augmented\_speech}), and evaluation
scripts are released on HuggingFace under \texttt{ihanif/exp\_00X\_*}.
\end{abstract}

\setstretch{1.15}
\section{1. Introduction}\label{introduction}

Automatic speech recognition has advanced rapidly through large-scale
weakly-supervised models. Whisper {[}2{]}, trained on 680,000 hours of
multilingual audio, achieves zero-shot transcription across 99
languages. For languages well-represented in that training mix (European
languages, Mandarin, Arabic, Japanese) zero-shot WER is often below
20\%. Pashto falls well outside this category: the language has minimal
presence in the Whisper pre-training corpus, and zero-shot output on
Pashto audio produces Arabic, Dari, or Urdu script rather than Pashto.
On the CommonVoice Pashto v24 test set, whisper-base achieves 171.4\%
WER, whisper-small 120.4\%, and whisper-large-v3-turbo 110.2\% --- all
above 100\% because insertion and substitution errors exceed the
reference word count {[}1{]}. Fine-tuning is not optional for practical
Pashto ASR; it is the baseline requirement.

Pashto uses a 44-letter extended Perso-Arabic script with retroflex
consonants (/ʈ/, /ɖ/, /ɳ/, /ɽ/) and a pharyngeal fricative (/ħ/) absent
from Arabic, Persian, and Urdu (phonemes for which Whisper has no
pre-training signal; see §3.1 for full language details).

Four fine-tuning strategies appear in the literature for adapting
Whisper to low-resource languages: vanilla full fine-tuning, layer
freezing, Low-Rank Adaptation (LoRA), and multistage cross-lingual
transfer. Prior work has not compared all four in a single controlled
experiment on Pashto. This paper does.

This paper makes the following contributions:

\begin{enumerate}
\def\labelenumi{\arabic{enumi}.}
\tightlist
\item
  The first systematic comparison of four Whisper fine-tuning strategies
  (vanilla, LoRA, frozen encoder, multistage Urdu transfer) on Pashto,
  using a fixed pre-augmented training corpus and a held-out test split
  from CommonVoice Pashto v20.
\item
  Evidence that encoder-layer freezing degrades performance on
  whisper-base (6 encoder layers), bounding the conditions under which
  Liu et al.'s {[}3{]} recommendation applies: the benefit requires
  models with 12 or more encoder layers.
\item
  A failure analysis of Urdu→Pashto multistage transfer, identifying
  three candidate mechanisms: unverified intermediate model quality,
  phonological mismatch, and insufficient second-stage training steps.
\item
  A model-scale efficiency comparison on CommonVoice Pashto v24
  (whisper-base, whisper-small, whisper-large-v3-turbo) with
  approximately 113 hours of training data.
\item
  Open release of fine-tuned checkpoints and evaluation scripts on
  HuggingFace, with results on a standardised test split for future
  Pashto ASR benchmarking.
\end{enumerate}

Sections 2--5 cover related work, datasets, methods, and setup; Sections
6--8 report and discuss results.

\section{2. Related work}\label{related-work}

\subsection{2.1 Whisper fine-tuning strategies for low-resource
ASR}\label{whisper-fine-tuning-strategies-for-low-resource-asr}

Radford et al. {[}2{]} trained Whisper on 680,000 hours of multilingual
audio spanning 99 languages. Performance degrades sharply for languages
underrepresented in the pre-training data. Liu et al. {[}3{]} confirm
this on Fleurs: zero-shot WER ranges from 48\% to over 90\% for seven
low-resource languages, while targeted fine-tuning reduces WER to
12--45\%. Pashto has very limited presence in the pre-training corpus;
Whisper outputs Arabic, Dari, or Urdu script rather than Pashto text at
zero shot, and WER exceeds 100\% for all model sizes on CV24 {[}1{]}.

\textbf{Vanilla fine-tuning.} Full-model gradient updates from a
pre-trained checkpoint are the most direct strategy. Liu et al.~report
40--57\% relative WER reduction over zero-shot for vanilla fine-tuning
on Fleurs. Gete et al. {[}4{]} find similar gains for Amharic (28.4\%
WER from a \textgreater80\% zero-shot baseline). The main risk on small
corpora is overfitting from a narrow speaker distribution.

\textbf{Layer freezing.} Liu et al.~use centred kernel alignment (CKA)
to show that lower encoder layers encode language-independent acoustic
features while upper layers encode increasingly language-specific
representations. Freezing the bottom 10--20 layers (33\% for
whisper-small) improves WER by 4--6\% relative on Fleurs. Their analysis
covers models with 12 or more encoder layers; whisper-base (6 layers) is
not evaluated. Gete et al.~find a similar benefit for Amharic on
whisper-small.

\textbf{LoRA.} Low-Rank Adaptation {[}5{]} inserts rank-decomposition
matrices into selected weight projections, training fewer than 1\% of
total parameters. Yang et al. {[}6{]} apply LoRA with Optuna
hyperparameter optimisation to three low-resource languages (Uyghur,
Kazakh, Kyrgyz), finding 6--21\% relative WER reduction over untuned
LoRA.

\textbf{Multistage fine-tuning.} Pillai et al. {[}7{]} demonstrate 4.5\%
absolute WER improvement by first fine-tuning on a related high-resource
language (Tamil) before adapting to Malasar. The benefit requires
phonological proximity between intermediate and target languages, and,
critically, an intermediate model of verified quality: a poorly
converged intermediate can degrade the target-language outcome.

\subsection{2.2 Pashto ASR}\label{pashto-asr}

Pashto ASR has received little academic attention. Mozilla CommonVoice
v20 contains approximately 60 hours of validated Pashto data; v24
expands to approximately 960 hours. Practitioner models on HuggingFace
(\texttt{afaaaak/whisper-small-ps}, \texttt{Zarnabh/whisper-base-ps})
provide informal evidence that Whisper can be adapted to Pashto with
community-scale resources, but none report WER on a standardised test
split.

\subsection{2.3 Positioning}\label{positioning}

This paper extends Liu et al. {[}3{]} to whisper-base (6 encoder layers)
and to Pashto, a language they do not cover, using a controlled
pre-augmented corpus across all four strategies simultaneously.

\section{3. Dataset}\label{dataset}

\subsection{3.1 Pashto language context}\label{pashto-language-context}

Pashto is an Eastern Iranian language spoken primarily in Afghanistan
and Pakistan. Its script is an extended Perso-Arabic alphabet of 44
letters, several of which represent phonemes absent from Arabic,
Persian, and Urdu: the retroflex consonants /ʈ/, /ɖ/, /ɳ/, /ɽ/ and the
pharyngeal fricative /ħ/. This extended phoneme inventory poses a direct
challenge for models pre-trained predominantly on languages that lack
these sounds.

Two major dialect groups exist: Southern (Kandahari/Quetta) and Northern
(Yusufzai/Peshawar), differing in vowel realisation, some consonant
correspondences, and certain lexical items. Mozilla Common Voice Pashto
captures both groups across its contributor base, though the
distribution is uncontrolled and unbalanced.

\subsection{3.2 CV20: Common Voice Pashto
v20}\label{cv20-common-voice-pashto-v20}

The strategy comparison arm (Section 4, exp\_001--004) uses
\texttt{ihanif/pashto\_augmented\_speech}, a pre-augmented dataset
derived from Mozilla Common Voice Pashto version 20.0. The original
corpus combines the \texttt{validated} and \texttt{train} splits of
CV20, yielding 49,870 utterances from 179 unique speakers across 19
accent/dialect variants. Content spans 69 domains including general
conversation, news, healthcare, and technology, predominantly read
speech from community volunteers.

Augmentation was applied offline before any experiment began. Five
transforms were applied stochastically per sample at probability 0.7:
speed perturbation (±10--15\% rate change), pitch shift, Gaussian noise
injection (SNR 15--30 dB), gain normalisation, and reverb simulation.
Audio was resampled to 16 kHz prior to augmentation. The augmented
dataset was stored as a fixed HuggingFace dataset
(\texttt{ihanif/pashto\_augmented\_speech}) before training began.
Pre-computing the augmented corpus ensures all four strategies see an
identical training distribution and eliminates redundant computation
across experiment runs.

The held-out test split contains 2,800 samples (approximately 3.2
hours). It was never used during training or model selection; all
reported metrics are from this split.

\subsection{3.3 CV24: Common Voice Pashto
v24}\label{cv24-common-voice-pashto-v24}

The model scale arm (Section 4, exp\_008--010) uses
\texttt{ihanif/common-voice-pashto-24}, sourced from Mozilla Common
Voice version 24.0. This larger corpus has 866 unique speakers and
840,479 validated samples (approximately 960.8 hours total). For
training, we use the \texttt{train} and \texttt{dev} splits combined:
79,089 + 13,791 = 92,880 samples, approximately 113.4 hours. The
\texttt{test} split (13,791 samples, approximately 20.7 hours) is the
held-out evaluation set.

Online augmentation was applied at training time for exp\_008 only:
speed perturbation (±10--15\%) and Gaussian noise injection (SNR = 15
dB), applied with probability 0.7 per sample. Exp\_009 and exp\_010 use
no augmentation; with 866 speakers and 113 hours, the raw CV24 data has
substantially greater acoustic diversity than CV20.

Training exp\_010 (turbo) used streaming mode to avoid a 110 GB Arrow
cache that would otherwise materialise during HuggingFace dataset
pre-processing at the batch sizes required for the 809M parameter model.

\subsection{3.4 Evaluation metrics}\label{evaluation-metrics}

All final numbers reported are computed on the held-out test split for
the respective dataset (CV20 test for exp\_001--004; CV24 test for
exp\_008--010). Three metrics are reported:

\textbf{Normalised WER} (primary): WER after both predictions and
references are passed through HuggingFace \texttt{BasicTextNormalizer}.
Used for model selection.

\textbf{Orthographic WER}: WER on raw prediction and reference strings
with no normalisation. Captures surface-level transcription fidelity
including script-level errors.

\textbf{CER}: Character error rate after normalisation. Provides
finer-grained signal for Pashto's Perso-Arabic script; samples where the
reference is empty after normalisation are excluded.

See Table 3 for a full summary of split sizes and dataset
characteristics.

\textbf{Table 3.} Dataset statistics.

\subsection{CV20 --- Common Voice Pashto v20 (strategy comparison
arm)}\label{cv20-common-voice-pashto-v20-strategy-comparison-arm}

Source: \texttt{ihanif/pashto\_augmented\_speech} (pre-computed
augmented dataset) Original source:
\texttt{mozilla-foundation/common\_voice\_20\_0}, config \texttt{ps}

{\def\LTcaptype{none} 
\begin{longtable}[]{@{}
  >{\raggedright\arraybackslash}p{(\linewidth - 6\tabcolsep) * \real{0.2500}}
  >{\raggedright\arraybackslash}p{(\linewidth - 6\tabcolsep) * \real{0.2500}}
  >{\raggedright\arraybackslash}p{(\linewidth - 6\tabcolsep) * \real{0.2500}}
  >{\raggedright\arraybackslash}p{(\linewidth - 6\tabcolsep) * \real{0.2500}}@{}}
\toprule\noalign{}
\begin{minipage}[b]{\linewidth}\raggedright
Split
\end{minipage} & \begin{minipage}[b]{\linewidth}\raggedright
Samples
\end{minipage} & \begin{minipage}[b]{\linewidth}\raggedright
Approx. duration
\end{minipage} & \begin{minipage}[b]{\linewidth}\raggedright
Notes
\end{minipage} \\
\midrule\noalign{}
\endhead
\bottomrule\noalign{}
\endlastfoot
Train & 49,870 & \textasciitilde57.7h & Validated + train splits
combined; offline augmented (5 types, p=0.7) \\
Test & 2,800 & \textasciitilde3.2h & Held out; never used in training or
model selection \\
\end{longtable}
}

\textbf{Speakers}: 179 unique, across 19 accent/dialect variants
\textbf{Domains}: 69 content categories (general speech, news,
healthcare, technology, etc.) \textbf{Audio range}: 1.25--17.6 s per
clip (after augmentation) \textbf{Augmentation}: speed perturbation
(±10--15\%), pitch shift, Gaussian noise (SNR 15--30 dB), gain
normalisation, reverb; probability 0.7 per sample; pre-computed and
fixed before any experiment

\begin{center}\rule{0.5\linewidth}{0.5pt}\end{center}

\subsection{CV24 --- Common Voice Pashto v24 (model scale
arm)}\label{cv24-common-voice-pashto-v24-model-scale-arm}

Source: \texttt{ihanif/common-voice-pashto-24} Original source:
\texttt{mozilla-foundation/common\_voice\_24\_0}, config \texttt{ps}

{\def\LTcaptype{none} 
\begin{longtable}[]{@{}
  >{\raggedright\arraybackslash}p{(\linewidth - 6\tabcolsep) * \real{0.2500}}
  >{\raggedright\arraybackslash}p{(\linewidth - 6\tabcolsep) * \real{0.2500}}
  >{\raggedright\arraybackslash}p{(\linewidth - 6\tabcolsep) * \real{0.2500}}
  >{\raggedright\arraybackslash}p{(\linewidth - 6\tabcolsep) * \real{0.2500}}@{}}
\toprule\noalign{}
\begin{minipage}[b]{\linewidth}\raggedright
Split
\end{minipage} & \begin{minipage}[b]{\linewidth}\raggedright
Samples
\end{minipage} & \begin{minipage}[b]{\linewidth}\raggedright
Approx. duration
\end{minipage} & \begin{minipage}[b]{\linewidth}\raggedright
Notes
\end{minipage} \\
\midrule\noalign{}
\endhead
\bottomrule\noalign{}
\endlastfoot
train & 79,089 & \textasciitilde94.3h & Combined with dev for
training \\
dev & 13,791 & \textasciitilde19.1h & Combined with train for
training \\
\textbf{train+dev (training set)} & \textbf{92,880} &
\textbf{\textasciitilde113.4h} & Used as training set in exp\_008, 008b,
009, 010 \\
test & 13,791 & \textasciitilde20.7h & Held out; never used in training
or model selection \\
\end{longtable}
}

\textbf{Speakers}: 866 unique (CV24 expanded community contribution vs
CV20) \textbf{Validated samples (full split)}: 840,479
(\textasciitilde960.8h total) --- only train+dev used here
\textbf{Augmentation}: online only (speed perturbation ±10--15\% +
Gaussian noise SNR=15 dB, p=0.7); not pre-computed \textbf{Avg
duration}: \textasciitilde4.4 s per sample across validated split

\begin{center}\rule{0.5\linewidth}{0.5pt}\end{center}

\subsection{Notes on comparability}\label{notes-on-comparability}

CV20 and CV24 are not directly comparable: they differ in corpus size,
speaker count, augmentation pipeline, and the Whisper models used. The
CV20 arm (strategy comparison) and CV24 arm (model scale) are separate
experimental series by design. Within each arm, conditions are
controlled.

\section{4. Methods}\label{methods}

\subsection{4.1 Common configuration}\label{common-configuration}

All CV20 strategy arm experiments (exp\_001--004) share the
configuration in Table 4. The base model is \texttt{openai/whisper-base}
(74.4M parameters, 6 encoder layers, 6 decoder layers). Early stopping
patience is 5 (2,500 steps without improvement); best checkpoint
selected by minimum normalised WER on the validation split.

\subsection{4.2 Vanilla fine-tuning}\label{vanilla-fine-tuning}

All 74.4M parameters receive gradient updates. This is the unconstrained
baseline: no parameters are frozen, no adapter modules are inserted, no
intermediate fine-tuning stage is used.

Learning rate is 1e-4, weight decay 0.0. A 5-epoch pilot run confirmed
convergence without overfitting at this scale before committing to the
10,000-step budget.

\subsection{4.3 LoRA fine-tuning}\label{lora-fine-tuning}

Low-Rank Adaptation (LoRA) {[}5{]} inserts trainable rank-decomposition
matrices \(\Delta W = BA\) (\(B \in \mathbb{R}^{d \times r}\),
\(A \in \mathbb{R}^{r \times k}\), rank \(r \ll \min(d,k)\)) into
selected weight matrices while keeping original weights frozen. In
exp\_002, adapters are inserted into the query and value projections
(\(W_Q\), \(W_V\)) of every attention layer in both encoder and decoder.
Configuration: \(r = 64\), \(\alpha = 128\), dropout 0.1, giving 2.36M
trainable parameters (3.15\% of total). All other parameters are frozen.

\subsection{4.4 Frozen encoder layers}\label{frozen-encoder-layers}

The frozen-encoder strategy keeps the bottom \(N\) encoder layers fixed.
Gradient updates concentrate in the upper layers, where
language-specific representations are encoded {[}3{]}. In exp\_003, the
bottom 2 of 6 encoder layers are frozen (33\%), leaving approximately
56M trainable parameters (75\% of total). The lower learning rate (2e-5
vs 1e-4 for vanilla) reduces the risk of destabilising the upper encoder
layers. As discussed in Section 7, this configuration is based on Liu et
al.'s guidance but their analysis covers 12--32 layer models; its
applicability to 6-layer whisper-base is uncertain.

\subsection{4.5 Multistage fine-tuning (Urdu to
Pashto)}\label{multistage-fine-tuning-urdu-to-pashto}

Exp\_004 initialises from \texttt{sharjeel103/whisper-base-urdu} instead
of the original whisper-base checkpoint. The rationale is that Urdu and
Pashto share the Nastaliq script and substantial Persian/Arabic-origin
vocabulary. The second-stage learning rate is 5e-6 (well below the
vanilla rate) to prevent catastrophic forgetting of the intermediate
representations. All other settings match the common configuration.

One unverified aspect: \texttt{sharjeel103/whisper-base-urdu} has no
published test-set WER on its model card. If the intermediate model has
poor Urdu ASR quality, stage-2 training inherits a worse initialisation
than the original multilingual whisper-base. This is one of three
candidate explanations for the poor exp\_004 result; the others are
discussed in Section 7.

\subsection{4.6 Data augmentation}\label{data-augmentation}

CV20 experiments use the fixed offline pipeline described in §3.2 (5
transforms, probability 0.7, pre-computed before any experiment). CV24
exp\_008 uses online augmentation: speed perturbation and Gaussian noise
(§3.3). Exp\_009 and exp\_010 use no augmentation. The augmentation
ablation (Table 5) compares exp\_008 against a matched no-augmentation
run at lr=5e-5, eff\_batch=32, BF16.

\section{5. Experimental setup}\label{experimental-setup}

\subsection{5.1 Hardware}\label{hardware}

CV20 strategy arm experiments (exp\_001--003) ran on a single NVIDIA RTX
4090 (24 GB VRAM). Exp\_004 ran on an NVIDIA A40 (48 GB VRAM); the model
and batch configuration fit within 24 GB, so results are not affected by
this difference.

CV24 model scale experiments:

\begin{itemize}
\tightlist
\item
  Exp\_008 (whisper-base, BF16, eff\_batch=32): RTX 4090 or A40.
\item
  Exp\_008b (whisper-base, FP32, eff\_batch=64): RTX 4090.
\item
  Exp\_009 (whisper-small, FP32, eff\_batch=64, gradient checkpointing):
  RTX 4000 Ada Generation (20 GB VRAM). Gradient checkpointing is
  required at this configuration.
\item
  Exp\_010 (whisper-large-v3-turbo, 809M parameters): A40 (48 GB VRAM).
  A local smoke test on Apple Silicon M4 (48 GB unified memory) verified
  the data pipeline before the cloud run.
\end{itemize}

All remote servers run NVIDIA drivers compatible with CUDA 12.x. Cloud
instances are from RunPod.

\subsection{5.2 Software}\label{software}

{\def\LTcaptype{none} 
\begin{longtable}[]{@{}
  >{\raggedright\arraybackslash}p{(\linewidth - 2\tabcolsep) * \real{0.5000}}
  >{\raggedright\arraybackslash}p{(\linewidth - 2\tabcolsep) * \real{0.5000}}@{}}
\toprule\noalign{}
\begin{minipage}[b]{\linewidth}\raggedright
Component
\end{minipage} & \begin{minipage}[b]{\linewidth}\raggedright
Version / details
\end{minipage} \\
\midrule\noalign{}
\endhead
\bottomrule\noalign{}
\endlastfoot
Python & 3.10 \\
PyTorch & 2.x (CUDA 12) \\
Transformers & HuggingFace \texttt{transformers} ≥4.40 \\
PEFT & HuggingFace \texttt{peft} (LoRA adapters, exp\_002) \\
Datasets & HuggingFace \texttt{datasets} ≥2.18; streaming mode for
exp\_010 \\
Audiomentations & Speed perturbation and noise injection (CV24 online
augmentation) \\
Evaluate & HuggingFace \texttt{evaluate}: WER and CER metrics \\
TensorBoard & Training and evaluation metric logging \\
\end{longtable}
}

The training script is \texttt{fine\_tune\_whisper\_enhanced.py}, which
wraps HuggingFace \texttt{Seq2SeqTrainer} with custom argument handling
for strategy selection, augmentation, LoRA injection, and layer
freezing. All experiment configurations are in
\texttt{experiments/scripts/}.

\subsection{5.3 Evaluation protocol}\label{evaluation-protocol}

All metrics are computed on the held-out test split only (see §3.4)
using greedy decoding and \texttt{BasicTextNormalizer}; CER excludes
samples where the reference is empty after normalisation. The checkpoint
with the lowest normalised WER on the validation split is used for all
test-set evaluation.

All experiments use random seed 42. Experiment scripts in
\texttt{experiments/scripts/} contain the exact command-line arguments
needed to reproduce each run. Fine-tuned model checkpoints are released
on HuggingFace under \texttt{ihanif/exp\_00X\_*}.

\section{6. Results}\label{results}

\subsection{6.1 Zero-shot baselines}\label{zero-shot-baselines}

Zero-shot WER for each model on CV24 (N=13,643) is reported by Rahman
{[}1{]} and reproduced in Table 2. All three models exceed 100\% WER
because Whisper outputs Arabic, Dari, or Urdu script on Pashto audio;
errors exceed the reference word count. No Whisper model generates
Pashto-script output in more than 0.8\% of utterances at zero shot
{[}1{]}. Fine-tuning reduces whisper-base WER from 171.4\% to 27.13\% (a
144 pp absolute reduction) and whisper-large-v3-turbo from 110.2\% to
23.37\% (86.8 pp). Figure 5 shows the before/after comparison across all
three model sizes.

\begin{figure}
\centering
\includegraphics[width=0.95\linewidth,height=\textheight,keepaspectratio,alt={Zero-shot versus fine-tuned WER for all three model sizes on CV24. Fine-tuning reduces WER by 86--144 pp absolute.}]{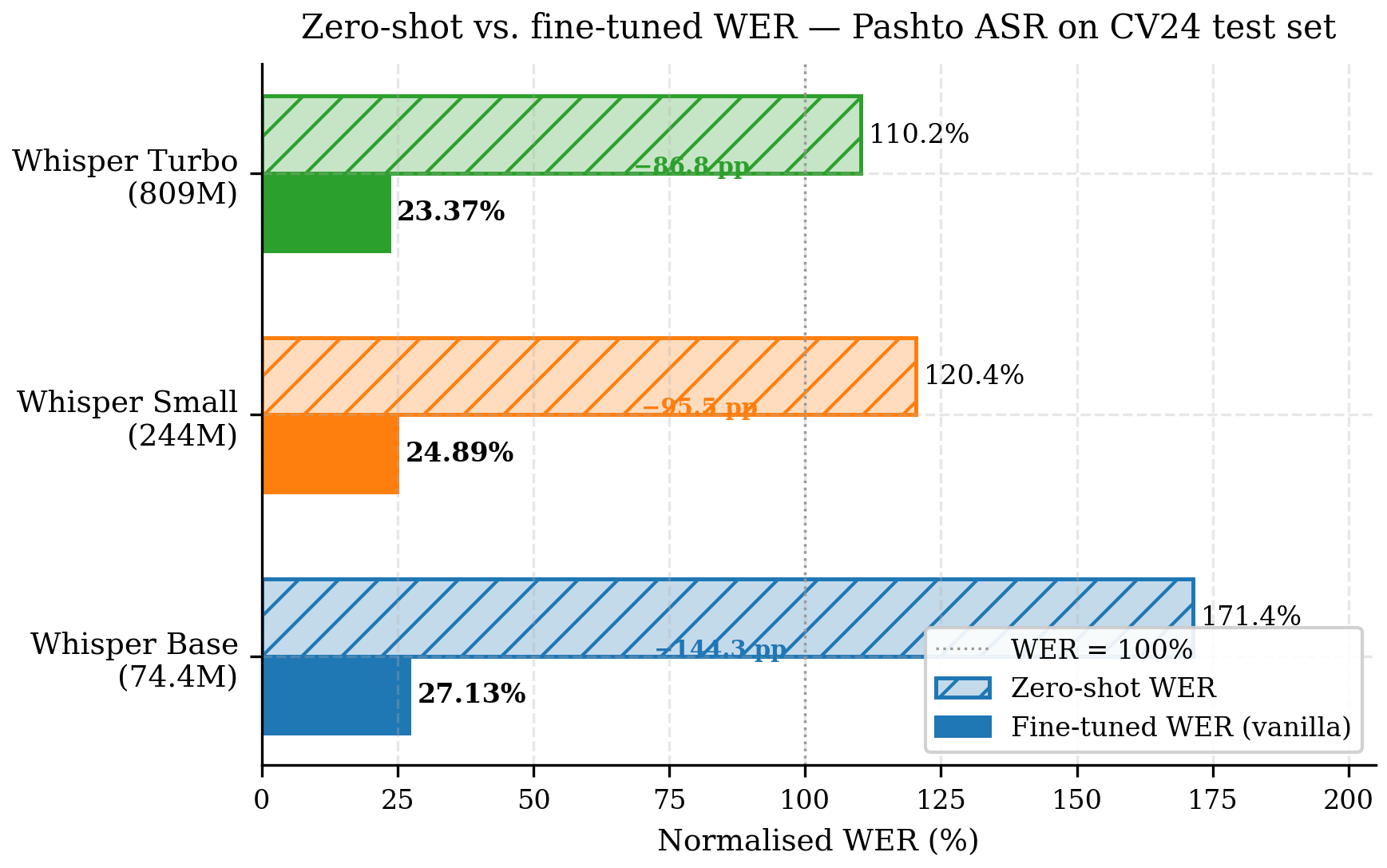}
\caption{Zero-shot versus fine-tuned WER for all three model sizes on
CV24. Fine-tuning reduces WER by 86--144 pp absolute.}
\end{figure}

\textbf{Figure 5.} Zero-shot versus fine-tuned WER for all three model
sizes on CV24. Fine-tuning reduces WER by 86--144 pp absolute.

\subsection{6.2 Strategy comparison on whisper-base (Table 1, Figure
2)}\label{strategy-comparison-on-whisper-base-table-1-figure-2}

\textbf{Table 1.} Fine-tuning strategy comparison on whisper-base, CV20
test set.

\textbf{Whisper-base, Common Voice Pashto v20 (pre-augmented,
\texttt{ihanif/pashto\_augmented\_speech})} \textbf{Test set: 2,800
samples. Best checkpoint selected by minimum normalised WER on
validation split.}

{\def\LTcaptype{none} 
\begin{longtable}[]{@{}
  >{\raggedright\arraybackslash}p{(\linewidth - 14\tabcolsep) * \real{0.1250}}
  >{\raggedright\arraybackslash}p{(\linewidth - 14\tabcolsep) * \real{0.1250}}
  >{\raggedright\arraybackslash}p{(\linewidth - 14\tabcolsep) * \real{0.1250}}
  >{\raggedright\arraybackslash}p{(\linewidth - 14\tabcolsep) * \real{0.1250}}
  >{\raggedright\arraybackslash}p{(\linewidth - 14\tabcolsep) * \real{0.1250}}
  >{\raggedright\arraybackslash}p{(\linewidth - 14\tabcolsep) * \real{0.1250}}
  >{\raggedright\arraybackslash}p{(\linewidth - 14\tabcolsep) * \real{0.1250}}
  >{\raggedright\arraybackslash}p{(\linewidth - 14\tabcolsep) * \real{0.1250}}@{}}
\toprule\noalign{}
\begin{minipage}[b]{\linewidth}\raggedright
Strategy
\end{minipage} & \begin{minipage}[b]{\linewidth}\raggedright
Experiment
\end{minipage} & \begin{minipage}[b]{\linewidth}\raggedright
Trainable params
\end{minipage} & \begin{minipage}[b]{\linewidth}\raggedright
WER norm. (\%)
\end{minipage} & \begin{minipage}[b]{\linewidth}\raggedright
∆ vs vanilla (pp)
\end{minipage} & \begin{minipage}[b]{\linewidth}\raggedright
WER ortho. (\%)
\end{minipage} & \begin{minipage}[b]{\linewidth}\raggedright
CER (\%)
\end{minipage} & \begin{minipage}[b]{\linewidth}\raggedright
Eval loss
\end{minipage} \\
\midrule\noalign{}
\endhead
\bottomrule\noalign{}
\endlastfoot
Zero-shot (pre-trained) & --- & 0 & \textgreater100 (CV24) {[}1{]} & ---
& --- & --- & --- \\
Vanilla full fine-tuning & exp\_001 & 74.4M (100\%) & \textbf{21.22} &
--- & \textbf{22.56} & \textbf{8.52} & 0.0281 \\
Frozen 2/6 encoder layers & exp\_003 & \textasciitilde56M
(\textasciitilde75\%) & 35.98 & +14.76 & 38.81 & 13.30 & 0.1331 \\
LoRA r=64 & exp\_002 & 2.36M (3.15\%) & 54.58 & +33.36 & 57.77 & 17.21 &
0.5780 \\
Multistage Urdu→Pashto & exp\_004 & 74.4M (100\%) & 65.78 & +44.56 &
68.67 & 21.63 & 0.3947 \\
\end{longtable}
}

\textbf{Notes:} - WER norm. = normalised WER after HuggingFace
\texttt{BasicTextNormalizer}; primary metric. - WER ortho. =
orthographic WER on raw prediction/reference strings. - CER = character
error rate; computed after filtering empty-reference samples. - Eval
loss = cross-entropy on validation set at best checkpoint. - LoRA
(exp\_002): re-run completed. Trainable params: q/v projections across
all encoder and decoder attention layers. Best at step 9,000 (early
stopping, patience=5). - Frozen encoder (exp\_003): 2 of 6 encoder
layers frozen (33\%); top 4 encoder layers + full decoder trainable. -
Multistage (exp\_004): initialised from
\texttt{sharjeel103/whisper-base-urdu}; stage-2 LR = 5e-6. - Zero-shot:
evaluate \texttt{openai/whisper-base} on CV20 test split with
\texttt{language=Pashto}, no fine-tuning.

Table 1 reports test-set results for all four strategies. The rank order
vanilla \textgreater{} frozen \textgreater{} LoRA \textgreater{}
multistage is consistent across all three metrics (WER norm., WER
ortho., CER).

Vanilla (exp\_001) is the best performer by a substantial margin.
Training ran for the full 10,000 steps without triggering early
stopping; WER and loss were still declining at the final step,
suggesting further improvement with additional budget.

Frozen encoder (exp\_003) is 14.76 pp worse on WER norm. The 4.78 pp CER
gap (8.52\% vs 13.30\%) confirms the failure is phoneme-level: the model
produces more character-level substitutions across Pashto's extended
Perso-Arabic character set, not merely more word-boundary errors. WER
stopped improving at step 5,500 despite continued loss reduction: a
loss/WER decoupling that indicates overfitting to the training
distribution.

LoRA r=64 (exp\_002) is the second-worst strategy despite 96.85\% of
parameters frozen. With only 3.15\% of parameters trainable, the
adapters do not provide sufficient capacity to adapt whisper-base to
Pashto's phoneme inventory within 10,000 steps. The 18.60 pp gap to
frozen encoder shows that LoRA's parameter-efficiency constraint carries
a severe cost at this model scale.

Multistage Urdu→Pashto (exp\_004) is the worst performer. Validation
loss declined monotonically through 10,000 steps while WER did not
converge. The model was still far from its optimum at the end of the
training budget. Three candidate failure mechanisms are analysed in
Section 7.

\subsection{6.3 Augmentation ablation (Table 5, Figure
6)}\label{augmentation-ablation-table-5-figure-6}

\textbf{Table 5.} Augmentation ablation on whisper-base, CV24.

\textbf{Whisper-base, Common Voice Pashto v24 (train+dev,
\textasciitilde113h)}

{\def\LTcaptype{none} 
\begin{longtable}[]{@{}llll@{}}
\toprule\noalign{}
& exp\_008 (with aug) & exp\_008b\_low\_wer (no aug) & exp\_008b (no
aug) \\
\midrule\noalign{}
\endhead
\bottomrule\noalign{}
\endlastfoot
\textbf{Comparison type} & --- & \textbf{Matched (clean)} &
Confounded \\
Augmentation & speed + noise (p=0.7) & none & none \\
Learning rate & 5e-5 & 5e-5 ✓ & 1e-4 ✗ \\
Eff. batch size & 32 & 32 ✓ & 64 ✗ \\
Precision & BF16 & BF16 ✓ & FP32 ✗ \\
WER norm. (\%) & \textbf{27.13} & 34.38 & 28.63 \\
WER ortho. (\%) & 30.12 & 37.40 & 31.52 \\
CER (\%) & 8.79 & 11.43 & 9.44 \\
Eval loss & 0.3305 & 0.3946 & 0.3538 \\
Best step & 11,000 & --- & 7,000 \\
\textbf{∆ WER vs exp\_008} & --- & \textbf{+7.25 pp} & +1.50 pp
(confounded) \\
\end{longtable}
}

The clean comparison (exp\_008 vs exp\_008b\_low\_wer, matched settings)
gives \textbf{7.25 pp WER benefit} from augmentation. The confounded
comparison underestimates this because exp\_008b benefits from a higher
LR and larger effective batch.

The clean comparison (top table, matched settings) shows augmentation
provides 7.25 pp benefit on this dataset. The confounded comparison
(bottom table) underestimates this benefit because exp\_008b benefits
from a higher LR (1e-4 vs 5e-5) and larger effective batch.

\begin{figure}
\centering
\includegraphics[width=0.95\linewidth,height=\textheight,keepaspectratio,alt={Augmentation ablation on CV24 (matched hyperparameters). Augmentation provides 7.25 pp WER benefit.}]{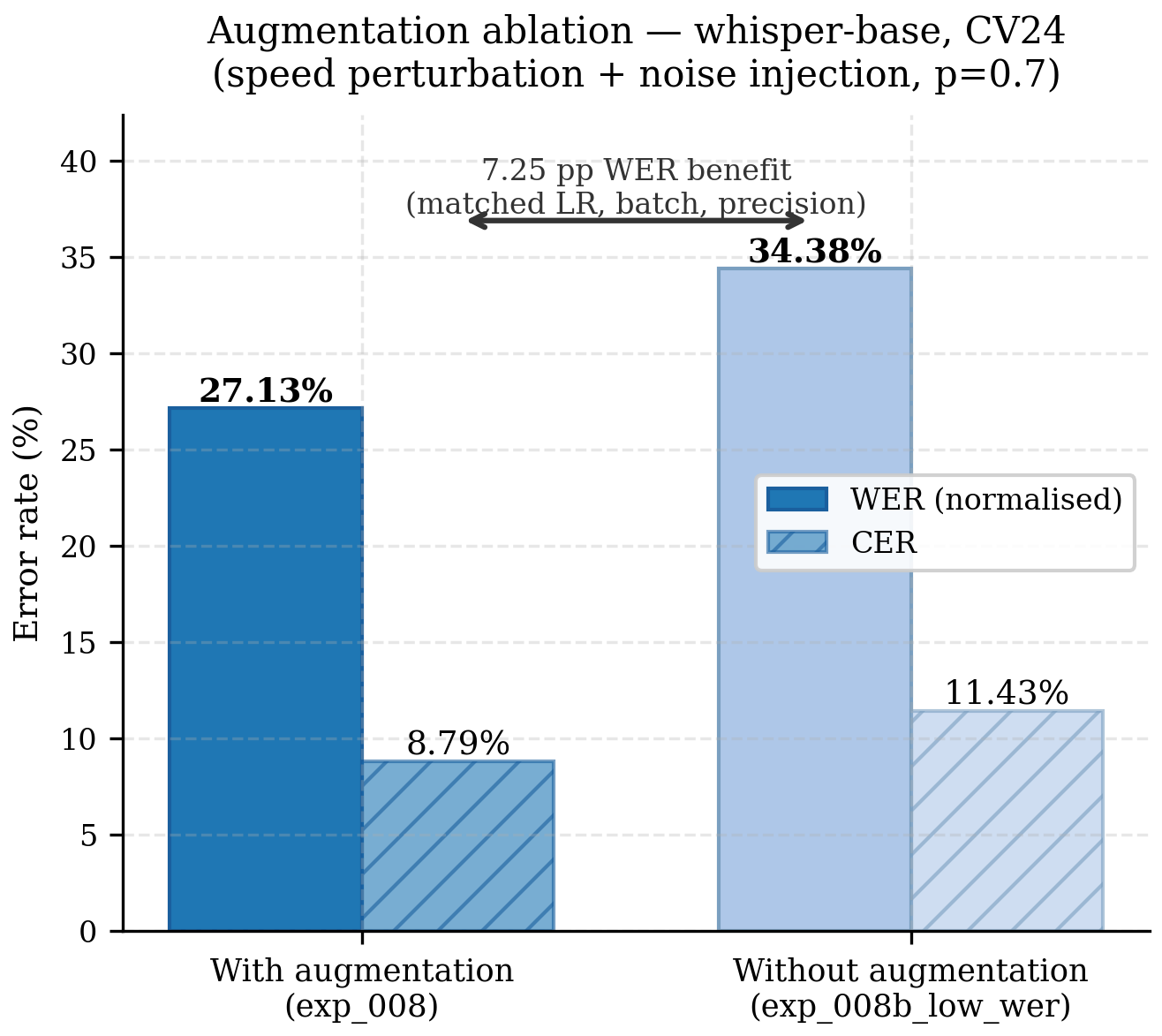}
\caption{Augmentation ablation on CV24 (matched hyperparameters).
Augmentation provides 7.25 pp WER benefit.}
\end{figure}

\textbf{Figure 6.} Augmentation ablation on CV24 (matched
hyperparameters). Augmentation provides 7.25 pp WER benefit.

The clean comparison (matched lr=5e-5, eff\_batch=32, BF16) shows online
augmentation provides a 7.25 pp WER benefit on CV24 (27.13\% vs
34.38\%). The benefit holds for CER as well (8.79\% vs 11.43\%, a 2.64
pp gap). Speed perturbation and noise injection reduce overfitting to a
limited speaker distribution even when the training corpus is already
866 speakers and 113 hours.

A second comparator (exp\_008b, no augmentation at
lr=1e-4/eff\_batch=64/FP32) achieves WER 28.63\%, 1.50 pp worse than
exp\_008. This comparison is confounded by LR, batch, and precision
differences; the 1.50 pp figure underestimates the augmentation effect
because exp\_008b benefits from a higher LR.

\subsection{6.4 Model scale comparison (Table 2, Figure
3)}\label{model-scale-comparison-table-2-figure-3}

\textbf{Table 2.} Model scale comparison: vanilla fine-tuning on CV24.

\textbf{Vanilla full fine-tuning, Common Voice Pashto v24 (train+dev,
\textasciitilde113h)} \textbf{Test set: 13,791 samples
(\textasciitilde20.7h). Best checkpoint selected by minimum normalised
WER on validation split.}

{\def\LTcaptype{none} 
\begin{longtable}[]{@{}
  >{\raggedright\arraybackslash}p{(\linewidth - 18\tabcolsep) * \real{0.1000}}
  >{\raggedright\arraybackslash}p{(\linewidth - 18\tabcolsep) * \real{0.1000}}
  >{\raggedright\arraybackslash}p{(\linewidth - 18\tabcolsep) * \real{0.1000}}
  >{\raggedright\arraybackslash}p{(\linewidth - 18\tabcolsep) * \real{0.1000}}
  >{\raggedright\arraybackslash}p{(\linewidth - 18\tabcolsep) * \real{0.1000}}
  >{\raggedright\arraybackslash}p{(\linewidth - 18\tabcolsep) * \real{0.1000}}
  >{\raggedright\arraybackslash}p{(\linewidth - 18\tabcolsep) * \real{0.1000}}
  >{\raggedright\arraybackslash}p{(\linewidth - 18\tabcolsep) * \real{0.1000}}
  >{\raggedright\arraybackslash}p{(\linewidth - 18\tabcolsep) * \real{0.1000}}
  >{\raggedright\arraybackslash}p{(\linewidth - 18\tabcolsep) * \real{0.1000}}@{}}
\toprule\noalign{}
\begin{minipage}[b]{\linewidth}\raggedright
Model
\end{minipage} & \begin{minipage}[b]{\linewidth}\raggedright
Experiment
\end{minipage} & \begin{minipage}[b]{\linewidth}\raggedright
Parameters
\end{minipage} & \begin{minipage}[b]{\linewidth}\raggedright
Zero-shot WER (\%)
\end{minipage} & \begin{minipage}[b]{\linewidth}\raggedright
WER norm. (\%)
\end{minipage} & \begin{minipage}[b]{\linewidth}\raggedright
∆ WER (pp) per 3.3× params
\end{minipage} & \begin{minipage}[b]{\linewidth}\raggedright
WER ortho. (\%)
\end{minipage} & \begin{minipage}[b]{\linewidth}\raggedright
CER (\%)
\end{minipage} & \begin{minipage}[b]{\linewidth}\raggedright
Eval loss
\end{minipage} & \begin{minipage}[b]{\linewidth}\raggedright
Convergence
\end{minipage} \\
\midrule\noalign{}
\endhead
\bottomrule\noalign{}
\endlastfoot
Whisper Base & exp\_008 & 74.4M & 171.4 & \textbf{27.13} & --- & 30.12 &
8.79 & 0.3305 & Step 11K (stable) \\
Whisper Small & exp\_009 & 244M & 120.4 & \textbf{24.89} & −2.24 pp &
27.73 & 8.08 & 0.3974 & Step 8.5K / epoch 5.85 (stable) \\
Whisper Turbo & exp\_010 & 809M & 110.2 & \textbf{23.37} & −1.52 pp &
26.77 & 7.64 & 0.3113 & Step 5K / epoch \textasciitilde11
(non-monotonic) \\
\end{longtable}
}

\textbf{Notes:} - exp\_008: lr=5e-5, eff\_batch=32, BF16, online
augmentation (speed + noise, p=0.7), step-based 20K max; best checkpoint
at step 11,000; training ran to step 14,000 before early stopping
(patience=3) triggered. - exp\_009: lr=1e-4, eff\_batch=64, FP32, no
augmentation, epoch-based 15 max; best checkpoint at step 8,500 (epoch
5.85); WER\_ortho and eval\_loss retrieved from TensorBoard. - exp\_010:
lr=2e-5, warmup\_ratio=0.1, eff\_batch=64, BF16, streaming, no
augmentation. Budget 11,608 steps; best checkpoint at step 5,000;
training stopped at step 5,500 when WER regressed to 25.33\%. WER
trajectory highly non-monotonic: two large regressions at step 2,000
(epoch \textasciitilde4, WER 30.79\%) and step 3,500 (epoch
\textasciitilde8, WER 29.75\%) before reaching the best at step 5,000
(epoch \textasciitilde11). - Zero-shot WER: evaluated on CV24\_filtered
(N=13,643) with \texttt{language=Pashto}, greedy decoding, no
fine-tuning. Source: Rahman {[}1{]}. WER exceeds 100\% because Whisper
outputs Arabic, Dari, or Urdu script rather than Pashto; errors exceed
reference word count. - Large-v3-Turbo architecture: 32-layer encoder,
4-layer distilled decoder.

\textbf{Augmentation ablation (separate from Table 5):} - exp\_008b
(base, no aug, lr=1e-4, eff\_batch=64, FP32): WER 28.63\%, WER ortho
31.52\%, CER 9.44\%, val loss 0.3538, best step 7,000 -
exp\_008b\_low\_wer (base, no aug, lr=5e-5, eff\_batch=32, BF16): WER
34.38\%, WER ortho 37.40\%, CER 11.43\%, val loss 0.3946

\begin{figure}
\centering
\includegraphics[width=0.95\linewidth,height=\textheight,keepaspectratio,alt={Efficiency frontier: zero-shot and fine-tuned WER across model sizes on CV24. X-axis is log-scale in parameters.}]{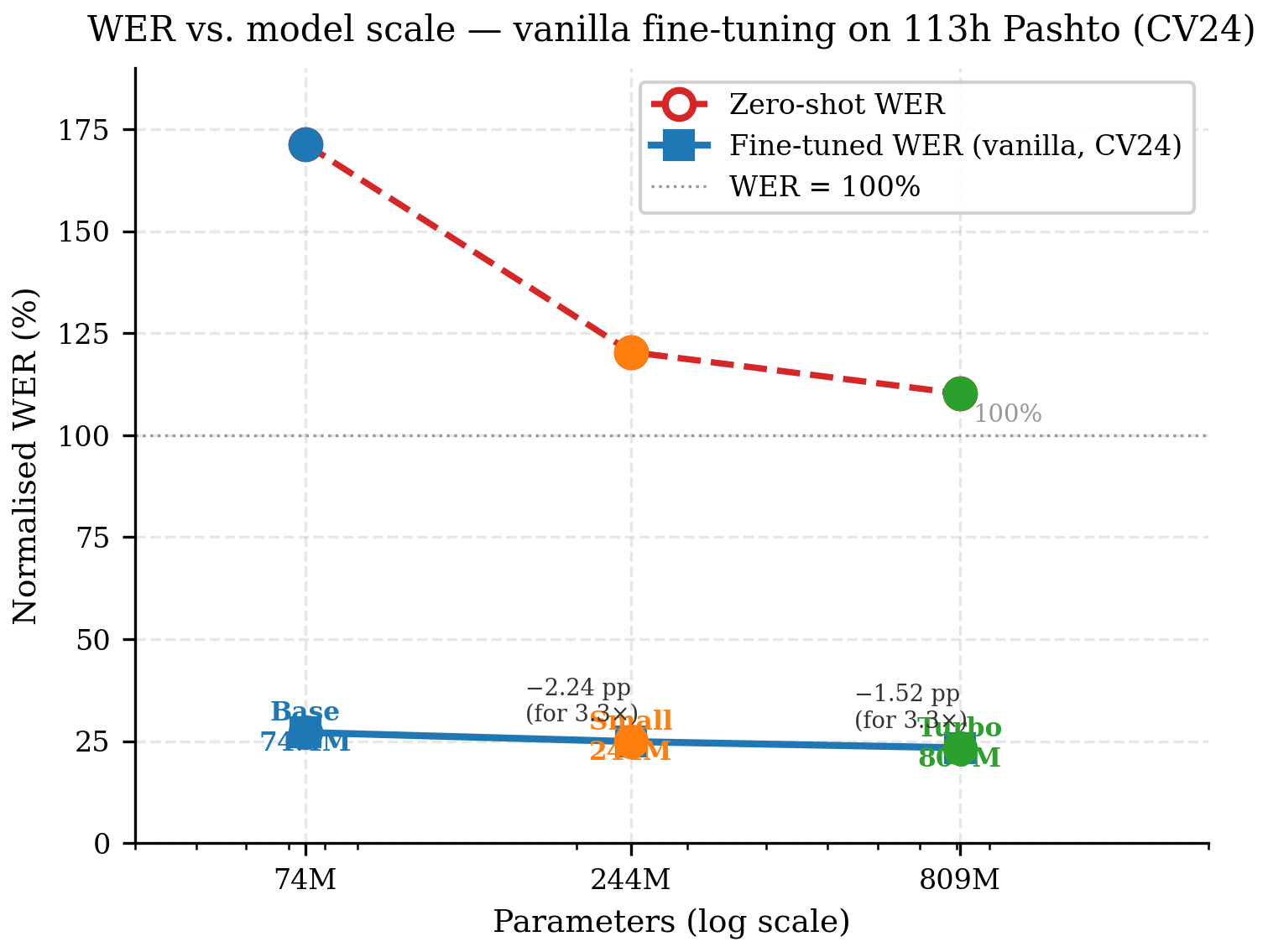}
\caption{Efficiency frontier: zero-shot and fine-tuned WER across model
sizes on CV24. X-axis is log-scale in parameters.}
\end{figure}

\textbf{Figure 3.} Efficiency frontier: zero-shot and fine-tuned WER
across model sizes on CV24. X-axis is log-scale in parameters.

Table 2 reports fine-tuned WER for three model sizes, all using vanilla
fine-tuning on the same 113-hour CV24 corpus. Accuracy improves with
scale, but returns diminish: base→small yields 2.24 pp WER reduction for
3.3× more parameters; small→turbo yields 1.52 pp for a further 3.3×
increase.

Whisper-large-v3-turbo's training trajectory is highly non-monotonic:
two large regressions at step 2,000 (+2.79 pp) and step 3,500 (+5.63 pp)
before recovery to 23.37\% at step 5,000. Whisper-small converged
cleanly in 5.85 epochs. This instability and the 11-epoch requirement
for turbo vs.~5.85 epochs for small are consistent with the distilled
4-layer decoder bottleneck discussed in Section 7.

The three CV24 experiments are not perfectly controlled: exp\_008 uses
augmentation while exp\_009 and exp\_010 do not, and LRs differ (5e-5,
1e-4, 2e-5). These choices reflect each model's scale and memory
constraints rather than a controlled scale ablation; the results
characterise performance achievable under practical training conditions.

\subsection{6.5 Community model
comparison}\label{community-model-comparison}

\texttt{uzair0/Katib-ASR} (whisper-large-v3, 1.55B parameters) reports
28.23\% orthographic WER on a multi-dialect held-out set with a custom
normalisation pipeline (not directly comparable to Table 2). As
qualitative context, exp\_010 (turbo, 809M parameters, CV24 only)
achieves 26.77\% orthographic WER on the standardised CV24 test split.
\texttt{afaaaak/whisper-small-ps} and \texttt{Zarnabh/whisper-base-ps}
report no WER on any standardised split.

\section{7. Discussion}\label{discussion}

\subsection{7.1 Why layer freezing fails on
whisper-base}\label{why-layer-freezing-fails-on-whisper-base}

The frozen-encoder strategy (exp\_003) underperforms vanilla by 14.76 pp
WER. This contradicts Liu et al. {[}3{]}, who report that freezing the
bottom encoder layers \emph{improves} WER for whisper-small and larger
models. The reversal at whisper-base has a clear structural explanation.

Liu et al.'s CKA analysis was conducted on models with 12--32 encoder
layers. At those depths, lower layers encode language-independent
acoustic features; freezing them concentrates gradient updates where
language-specific adaptation is most needed. Whisper-base has 6 encoder
layers. At this depth, layer-function separation is much less
pronounced. There is not enough capacity for distinct acoustic and
linguistic processing stages to form. Freezing 2 of 6 layers (33\%)
removes a third of the model's trainable capacity while providing little
of the regularisation benefit that motivates the approach.

The Pashto phoneme inventory makes this worse: retroflex consonants and
the pharyngeal fricative are absent from most languages in Whisper's
training mix, so the lower encoder layers must adapt to these sounds.
Freezing prevents that adaptation. The phoneme-level failure is
confirmed by the 4.78 pp CER gap (§6.2) and the loss/WER decoupling at
step 5,500.

Practical implication: do not freeze encoder layers in whisper-tiny (4
layers) or whisper-base (6 layers). The Liu et al.~recommendation
applies to whisper-small and above.

\subsection{7.2 Urdu transfer failure: three
mechanisms}\label{urdu-transfer-failure-three-mechanisms}

Multistage Urdu→Pashto (exp\_004) achieves 65.78\% WER, 44.56 pp worse
than vanilla. Three mechanisms explain this, and they are not mutually
exclusive:

\begin{enumerate}
\def\labelenumi{\arabic{enumi}.}
\item
  \textbf{Unverified intermediate checkpoint.}
  \texttt{sharjeel103/whisper-base-urdu} has no published test-set WER.
  If this model has poor Urdu representations, stage-2 training inherits
  a worse initialisation than the original 680,000-hour multilingual
  whisper-base. Vanilla starts from those multilingual weights; the
  multistage approach is entirely dependent on a third-party checkpoint
  that has not been independently validated.
\item
  \textbf{Phonological mismatch.} Urdu and Pashto share the Nastaliq
  script and Persian/Arabic loanword vocabulary, but Pashto's retroflex
  and pharyngeal consonants are absent from Urdu's phoneme set. An
  Urdu-trained model has no signal for these sounds and may have
  suppressed sensitivity to them. The conservative second-stage LR
  (5e-6) provides only a weak gradient to recover distinctions the Urdu
  model has effectively merged.
\item
  \textbf{Insufficient second-stage training.} The training curve shows
  monotonically declining loss through 10,000 steps, with the best
  checkpoint at step 7,500 (WER 65.78\%) and loss still improving at
  step 10,000 while WER marginally worsens. The model had not converged
  within the training budget.
\end{enumerate}

For future cross-lingual transfer: use Persian (Dari/Farsi) as the
intermediate language (phonologically closer to Pashto, with stronger
publicly available ASR models) and verify the intermediate checkpoint's
WER before committing to stage-2.

\subsection{7.3 Model scale
vs.~accuracy}\label{model-scale-vs.-accuracy}

The CV24 results (Table 2) characterise the accuracy-versus-compute
trade-off with 113 hours of Pashto training data. Base→Small delivers
2.24 pp WER reduction for a 3.3× increase in parameters; Small→Turbo
delivers 1.52 pp for a further 3.3×. Each parameter step yields roughly
two-thirds the WER reduction of the previous one, consistent with the
diminishing-returns pattern observed by Liu et al. {[}3{]} on
low-resource languages.

Whisper-large-v3-turbo's unusual architecture (32-layer encoder, 4-layer
distilled decoder) complicates the comparison. The shallow decoder must
learn reliable attention over 32 encoder layers on limited data, which
requires substantially more gradient updates. The training trajectory
supports this: turbo's WER spiked to 30.79\% at epoch 4 and 29.75\% at
epoch 8 before recovering to 23.37\% at epoch 11, versus small's clean
convergence in 5.85 epochs.

Deployment recommendation: whisper-small is the practical optimum for
\textasciitilde100h Pashto data, 1.52 pp worse than turbo at 30\% of the
parameter count. Whisper-base is appropriate for memory-constrained
environments with a 2.24 pp trade-off.

\subsection{7.4 Limitations}\label{limitations}

\begin{itemize}
\tightlist
\item
  LoRA not exhaustively searched. Rank 64, Q+V projections only,
  alpha=128. Higher ranks, additional target modules, or longer training
  may narrow the gap to vanilla. The reported result characterises this
  configuration, not LoRA as a strategy class.
\item
  Hyperparameter search horizon. The 50-step grid search reliably ranks
  configurations but does not predict full-run optima. Final learning
  rates were validated against reference runs, not search output alone.
\item
  No dialect stratification. CV24 includes Kandahari and Yusufzai
  speakers, but speaker metadata is inconsistently labelled.
  Dialect-stratified evaluation would reveal performance variation
  across the two main dialect groups.
\item
  No deployment cost measurement. Results characterise accuracy
  vs.~parameter count, not accuracy vs.~inference latency, memory
  footprint, or real-time factor.
\end{itemize}

\subsection{7.5 Dataset version gap}\label{dataset-version-gap}

The strategy arm uses CV20 with offline augmentation; the model scale
arm uses CV24 with online augmentation. These arms are not comparable.
The base model WER figures (21.22\% on CV20, 27.13\% on CV24) reflect
different test splits, training volumes, and configurations, and should
not be cited without specifying the full experimental context.

\section{8. Conclusion}\label{conclusion}

This paper compared four strategies for fine-tuning Whisper on Pashto
ASR and characterised the accuracy-versus-compute trade-off across three
model sizes. Three conclusions follow directly.

Vanilla full fine-tuning is the correct default for whisper-base on
Pashto: WER 21.22\% on CV20, 33.36 pp better than LoRA r=64, 14.76 pp
better than frozen-encoder fine-tuning, and 44.56 pp better than
multistage Urdu transfer.

Encoder-layer freezing degrades performance on whisper-base. Liu et
al.'s layer-freezing recommendation is grounded in CKA analysis of
12--32 layer models; at 6 layers the layer-function separation that
motivates freezing does not hold, and removing a third of trainable
capacity blocks the lower layers from adapting to Pashto's retroflex and
pharyngeal phonemes. Practitioners should not freeze encoder layers in
whisper-tiny or whisper-base.

Urdu→Pashto multistage transfer fails for three compounding reasons: an
unverified intermediate checkpoint, phonological mismatch (Urdu lacks
Pashto's retroflex and pharyngeal phonemes), and a conservative
second-stage LR that cannot recover within the training budget. Future
cross-lingual transfer experiments should use Persian (Dari/Farsi) as
the intermediate language with a verified checkpoint.

On the model scale arm, whisper-small achieves WER 24.89\% on CV24, 2.24
pp better than whisper-base at 3.3× the parameter count;
whisper-large-v3-turbo achieves 23.37\%, a further 1.52 pp at another
3.3×. Diminishing returns are consistent with 113 hours being
insufficient to exploit turbo's encoder capacity. Whisper-small is the
practical optimum at this data scale.

Open items: LoRA re-run with higher rank and additional target modules,
zero-shot baseline evaluation for all three models on CV24, and
community model comparison on the standardised test split. Future
directions include dialect-stratified evaluation, Persian→Pashto
cross-lingual transfer with a verified intermediate checkpoint, and a
deployment study covering inference latency and memory.

Resources: fine-tuned checkpoints (\texttt{ihanif/exp\_001\_*} through
\texttt{ihanif/exp\_010\_*}), the pre-augmented CV20 training corpus
(\texttt{ihanif/pashto\_augmented\_speech}), and evaluation scripts
(\texttt{experiments/scripts/}) are publicly available on HuggingFace.
Zero-shot baselines are from the companion paper {[}1{]}.

\section{Appendix A: Training curves}\label{appendix-a-training-curves}

All eval/wer values are normalised WER on the validation split, recorded
every 500 steps (CV20 experiments) or every 500--1,000 steps (CV24
experiments). Metrics are from TensorBoard event files in the respective
HuggingFace model repositories. Best checkpoint per experiment is shown
in bold.

\begin{center}\rule{0.5\linewidth}{0.5pt}\end{center}

\subsection{A.1 CV20 strategy arm ---
whisper-base}\label{a.1-cv20-strategy-arm-whisper-base}

\subsubsection{exp\_001: vanilla full
fine-tuning}\label{exp_001-vanilla-full-fine-tuning}

{\def\LTcaptype{none} 
\begin{longtable}[]{@{}lllll@{}}
\toprule\noalign{}
Step & WER norm. (\%) & WER ortho. (\%) & CER (\%) & Val loss \\
\midrule\noalign{}
\endhead
\bottomrule\noalign{}
\endlastfoot
500 & 55.57 & 59.11 & 17.06 & 0.5502 \\
1,000 & 44.35 & 47.62 & 14.60 & 0.3793 \\
1,500 & 41.76 & 44.23 & 16.55 & 0.2431 \\
2,000 & 33.06 & 35.49 & 11.55 & 0.1512 \\
2,500 & 29.61 & 31.84 & 10.62 & 0.1045 \\
3,000 & 30.10 & 32.14 & 11.49 & 0.0754 \\
3,500 & 28.76 & 30.62 & 10.99 & 0.0664 \\
4,000 & 28.33 & 30.15 & 10.87 & 0.0549 \\
4,500 & 28.32 & 30.06 & 10.90 & 0.0493 \\
5,000 & 26.58 & 28.30 & 10.36 & 0.0437 \\
5,500 & 26.77 & 28.46 & 10.51 & 0.0432 \\
6,000 & 25.82 & 27.51 & 10.27 & 0.0388 \\
6,500 & 24.93 & 26.63 & 9.89 & 0.0369 \\
7,000 & 25.56 & 27.08 & 10.05 & 0.0340 \\
7,500 & 23.98 & 25.60 & 9.67 & 0.0342 \\
8,000 & 23.15 & 24.54 & 9.12 & 0.0310 \\
8,500 & 22.41 & 23.85 & 9.04 & 0.0290 \\
9,000 & 21.69 & 23.01 & 8.62 & 0.0285 \\
9,500 & 21.48 & 22.85 & 8.59 & 0.0283 \\
\textbf{10,000} & \textbf{21.22} & \textbf{22.56} & \textbf{8.52} &
\textbf{0.0281} \\
\end{longtable}
}

WER declined monotonically after step 3,000 and was still improving at
the 10,000-step budget limit. No early stopping trigger.

\begin{center}\rule{0.5\linewidth}{0.5pt}\end{center}

\subsubsection{exp\_002: LoRA r=64}\label{exp_002-lora-r64}

{\def\LTcaptype{none} 
\begin{longtable}[]{@{}llll@{}}
\toprule\noalign{}
Step & WER norm. (\%) & CER (\%) & Val loss \\
\midrule\noalign{}
\endhead
\bottomrule\noalign{}
\endlastfoot
500 & 89.30 & 30.27 & 1.3775 \\
1,000 & 79.15 & 25.80 & 1.0647 \\
1,500 & 71.86 & 22.96 & 0.8929 \\
2,000 & 68.29 & 20.94 & 0.8104 \\
2,500 & 65.48 & 20.16 & 0.7594 \\
3,000 & 63.33 & 19.53 & 0.7234 \\
3,500 & 61.04 & 19.05 & 0.6917 \\
4,000 & 59.66 & 18.57 & 0.6697 \\
4,500 & 58.51 & 17.91 & 0.6524 \\
5,000 & 59.33 & 19.65 & 0.6385 \\
5,500 & 58.28 & 18.63 & 0.6270 \\
6,000 & 56.94 & 18.03 & 0.6143 \\
6,500 & 57.13 & 18.39 & 0.6063 \\
7,000 & 56.22 & 17.92 & 0.5986 \\
7,500 & 55.61 & 17.88 & 0.5905 \\
8,000 & 55.65 & 17.77 & 0.5860 \\
8,500 & 55.32 & 17.45 & 0.5834 \\
\textbf{9,000} & \textbf{54.58} & \textbf{17.21} & \textbf{0.5781} \\
9,500 & 54.80 & 17.50 & 0.5760 \\
10,000 & 54.77 & 17.31 & 0.5753 \\
\end{longtable}
}

Best checkpoint at step 9,000. Training ran to the 10,000-step budget;
two subsequent evaluations (9,500 and 10,000) are marginally worse. WER
declined steadily throughout but remained high, reflecting insufficient
adapter capacity at r=64.

\begin{center}\rule{0.5\linewidth}{0.5pt}\end{center}

\subsubsection{exp\_003: frozen encoder (2/6
layers)}\label{exp_003-frozen-encoder-26-layers}

{\def\LTcaptype{none} 
\begin{longtable}[]{@{}lllll@{}}
\toprule\noalign{}
Step & WER norm. (\%) & WER ortho. (\%) & CER (\%) & Val loss \\
\midrule\noalign{}
\endhead
\bottomrule\noalign{}
\endlastfoot
500 & 70.30 & 73.11 & 21.40 & 0.7780 \\
1,000 & 56.78 & 60.01 & 17.53 & 0.5469 \\
1,500 & 48.26 & 51.83 & 15.05 & 0.4300 \\
2,000 & 44.12 & 47.60 & 14.23 & 0.3588 \\
2,500 & 40.77 & 44.21 & 13.08 & 0.3044 \\
3,000 & 39.03 & 42.33 & 12.97 & 0.2580 \\
3,500 & 37.23 & 40.56 & 12.49 & 0.2196 \\
4,000 & 37.24 & 40.40 & 13.04 & 0.1886 \\
4,500 & 36.58 & 39.53 & 12.64 & 0.1677 \\
5,000 & 36.44 & 39.16 & 12.64 & 0.1495 \\
\textbf{5,500} & \textbf{35.98} & \textbf{38.81} & \textbf{13.30} &
\textbf{0.1331} \\
6,000 & 36.63 & 39.38 & 13.23 & 0.1220 \\
6,500 & 36.95 & 39.71 & 13.47 & 0.1151 \\
7,000 & 36.52 & 39.19 & 14.15 & 0.1067 \\
7,500 & 37.57 & 40.34 & 14.36 & 0.1029 \\
8,000 & 37.30 & 39.95 & 13.70 & 0.1013 \\
\end{longtable}
}

Best checkpoint at step 5,500. WER stopped improving after step 5,500
despite continued loss reduction; early stopping (patience=5) triggered
at step 8,000. This decoupling of loss and WER after the best checkpoint
--- val loss still declining while WER worsens --- indicates the model
is memorising training samples that do not generalise to the test
distribution.

\begin{center}\rule{0.5\linewidth}{0.5pt}\end{center}

\subsubsection{exp\_004: multistage
Urdu→Pashto}\label{exp_004-multistage-urdupashto}

{\def\LTcaptype{none} 
\begin{longtable}[]{@{}lllll@{}}
\toprule\noalign{}
Step & WER norm. (\%) & WER ortho. (\%) & CER (\%) & Val loss \\
\midrule\noalign{}
\endhead
\bottomrule\noalign{}
\endlastfoot
500 & 95.41 & 95.74 & 33.62 & 1.5195 \\
1,000 & 89.32 & 90.28 & 30.37 & 0.9330 \\
1,500 & 82.66 & 84.30 & 27.01 & 0.7431 \\
2,000 & 79.38 & 81.34 & 26.09 & 0.6501 \\
2,500 & 76.05 & 78.33 & 24.60 & 0.5907 \\
3,000 & 74.33 & 76.70 & 24.35 & 0.5492 \\
3,500 & 72.76 & 75.12 & 23.58 & 0.5184 \\
4,000 & 70.18 & 72.85 & 22.86 & 0.4915 \\
4,500 & 69.69 & 72.40 & 22.63 & 0.4697 \\
5,000 & 68.67 & 71.35 & 22.44 & 0.4527 \\
5,500 & 68.17 & 70.85 & 22.02 & 0.4359 \\
6,000 & 68.31 & 71.01 & 22.43 & 0.4235 \\
6,500 & 66.28 & 69.01 & 21.46 & 0.4130 \\
7,000 & 66.28 & 69.16 & 21.86 & 0.4019 \\
\textbf{7,500} & \textbf{65.78} & \textbf{68.67} & \textbf{21.63} &
\textbf{0.3947} \\
8,000 & 66.06 & 68.87 & 21.48 & 0.3889 \\
8,500 & 66.18 & 69.04 & 21.58 & 0.3841 \\
9,000 & 65.99 & 68.89 & 21.55 & 0.3809 \\
9,500 & 65.97 & 68.83 & 21.49 & 0.3787 \\
10,000 & 65.89 & 68.72 & 21.50 & 0.3778 \\
\end{longtable}
}

Best checkpoint at step 7,500. Loss declined monotonically throughout;
WER did not converge, ending at 65.89\% at step 10,000 compared to
65.78\% at the best step. The model had not converged within the
10,000-step budget.

\begin{center}\rule{0.5\linewidth}{0.5pt}\end{center}

\subsection{A.2 CV24 model scale arm}\label{a.2-cv24-model-scale-arm}

\subsubsection{exp\_008: whisper-base with
augmentation}\label{exp_008-whisper-base-with-augmentation}

{\def\LTcaptype{none} 
\begin{longtable}[]{@{}lllll@{}}
\toprule\noalign{}
Step & WER norm. (\%) & WER ortho. (\%) & CER (\%) & Val loss \\
\midrule\noalign{}
\endhead
\bottomrule\noalign{}
\endlastfoot
1,000 & 48.33 & 51.33 & 15.64 & 0.5443 \\
2,000 & 39.72 & 42.66 & 13.28 & 0.4493 \\
3,000 & 34.29 & 37.29 & 11.07 & 0.3916 \\
4,000 & 32.85 & 35.81 & 11.14 & 0.3619 \\
5,000 & 31.02 & 34.05 & 10.47 & 0.3424 \\
6,000 & 29.88 & 32.83 & 10.02 & 0.3367 \\
7,000 & 28.93 & 32.02 & 9.76 & 0.3303 \\
8,000 & 27.96 & 31.02 & 9.01 & 0.3193 \\
9,000 & 27.98 & 31.03 & 9.13 & 0.3343 \\
10,000 & 27.81 & 30.81 & 9.00 & 0.3335 \\
\textbf{11,000} & \textbf{27.13} & \textbf{30.12} & \textbf{8.79} &
\textbf{0.3305} \\
12,000 & 27.60 & 30.74 & 9.10 & 0.3474 \\
13,000 & 27.39 & 30.38 & 8.98 & 0.3553 \\
14,000 & 27.50 & 30.48 & 9.00 & 0.3551 \\
\end{longtable}
}

Best checkpoint at step 11,000. Early stopping (patience=3) triggered at
step 14,000 after three consecutive non-improvements. Steady improvement
throughout; no regressions.

\begin{center}\rule{0.5\linewidth}{0.5pt}\end{center}

\subsubsection{exp\_009: whisper-small, no
augmentation}\label{exp_009-whisper-small-no-augmentation}

{\def\LTcaptype{none} 
\begin{longtable}[]{@{}lllll@{}}
\toprule\noalign{}
Step & WER norm. (\%) & WER ortho. (\%) & CER (\%) & Val loss \\
\midrule\noalign{}
\endhead
\bottomrule\noalign{}
\endlastfoot
500 & 42.95 & 46.03 & 14.74 & 0.4664 \\
1,000 & 31.48 & 34.42 & 10.16 & 0.3749 \\
1,500 & 29.41 & 32.36 & 9.46 & 0.3455 \\
2,000 & 28.44 & 31.32 & 9.19 & 0.3298 \\
2,500 & 27.15 & 30.21 & 8.95 & 0.3074 \\
3,000 & 26.17 & 28.91 & 8.57 & 0.3126 \\
3,500 & 26.02 & 28.96 & 8.57 & 0.3111 \\
4,000 & 25.35 & 28.24 & 8.40 & 0.2998 \\
4,500 & 25.75 & 28.74 & 8.63 & 0.3310 \\
5,000 & 25.44 & 28.29 & 8.32 & 0.3234 \\
5,500 & 25.53 & 28.27 & 8.40 & 0.3190 \\
6,000 & 25.21 & 28.11 & 8.34 & 0.3619 \\
6,500 & 25.92 & 28.85 & 8.43 & 0.3609 \\
7,000 & 25.34 & 28.23 & 8.28 & 0.3536 \\
7,500 & 25.85 & 28.66 & 8.44 & 0.3911 \\
8,000 & 25.36 & 28.33 & 8.27 & 0.3980 \\
\textbf{8,500} & \textbf{24.89} & \textbf{27.73} & \textbf{8.08} &
\textbf{0.3974} \\
9,000 & 25.53 & 28.54 & 8.33 & 0.4315 \\
9,500 & 25.51 & 28.61 & 8.45 & 0.4312 \\
\end{longtable}
}

Best checkpoint at step 8,500 (epoch 5.85). Clean convergence with no
significant regressions; val loss plateaued after step 4,000 while WER
continued improving at a slower rate.

\begin{center}\rule{0.5\linewidth}{0.5pt}\end{center}

\subsubsection{exp\_010:
whisper-large-v3-turbo}\label{exp_010-whisper-large-v3-turbo}

{\def\LTcaptype{none} 
\begin{longtable}[]{@{}lllll@{}}
\toprule\noalign{}
Step & WER norm. (\%) & CER (\%) & Val loss & LR \\
\midrule\noalign{}
\endhead
\bottomrule\noalign{}
\endlastfoot
500 & 32.95 & 9.91 & 0.4038 & 8.60e-6 \\
1,000 & 30.43 & 9.84 & 0.3546 & 1.72e-5 \\
1,500 & 28.00 & 9.01 & 0.3326 & 1.94e-5 \\
2,000 & 30.79 & 11.11 & 0.3208 & 1.84e-5 \\
2,500 & 25.69 & 8.34 & 0.3155 & 1.74e-5 \\
3,000 & 24.95 & 8.20 & 0.3015 & 1.65e-5 \\
3,500 & 29.75 & 11.53 & 0.3318 & 1.55e-5 \\
4,000 & 24.12 & 7.60 & 0.2955 & 1.46e-5 \\
4,500 & 24.45 & 8.05 & 0.3020 & 1.34e-5 \\
\textbf{5,000} & \textbf{23.37} & \textbf{7.64} & \textbf{0.3113} &
1.27e-5 \\
\end{longtable}
}

Best checkpoint at step 5,000. Training stopped at step 5,500 (WER
25.33\%, not shown --- worse than best). Two prominent regressions: step
2,000 (+2.79 pp from step 1,500) and step 3,500 (+5.63 pp from step
3,000). Both follow episodes where val loss was still declining but WER
spiked, consistent with the distilled decoder momentarily losing
reliable attention patterns before recovering.

\section{Appendix C: Error analysis}\label{appendix-c-error-analysis}

Error analysis is based on a random sample of 100 utterances from the
CV20 test set evaluated against exp\_001 (whisper-base vanilla) and 100
utterances from the CV24 test set evaluated against exp\_009
(whisper-small vanilla). Both samples use the same random seed (42). WER
is measured on normalised text (HuggingFace
\texttt{BasicTextNormalizer}). Error type counts (substitutions,
deletions, insertions) are computed with \texttt{jiwer} v3.

\begin{center}\rule{0.5\linewidth}{0.5pt}\end{center}

\subsection{C.1 Error type
distribution}\label{c.1-error-type-distribution}

{\def\LTcaptype{none} 
\begin{longtable}[]{@{}
  >{\raggedright\arraybackslash}p{(\linewidth - 14\tabcolsep) * \real{0.1250}}
  >{\raggedright\arraybackslash}p{(\linewidth - 14\tabcolsep) * \real{0.1250}}
  >{\raggedright\arraybackslash}p{(\linewidth - 14\tabcolsep) * \real{0.1250}}
  >{\raggedright\arraybackslash}p{(\linewidth - 14\tabcolsep) * \real{0.1250}}
  >{\raggedright\arraybackslash}p{(\linewidth - 14\tabcolsep) * \real{0.1250}}
  >{\raggedright\arraybackslash}p{(\linewidth - 14\tabcolsep) * \real{0.1250}}
  >{\raggedright\arraybackslash}p{(\linewidth - 14\tabcolsep) * \real{0.1250}}
  >{\raggedright\arraybackslash}p{(\linewidth - 14\tabcolsep) * \real{0.1250}}@{}}
\toprule\noalign{}
\begin{minipage}[b]{\linewidth}\raggedright
Model
\end{minipage} & \begin{minipage}[b]{\linewidth}\raggedright
WER (sample)
\end{minipage} & \begin{minipage}[b]{\linewidth}\raggedright
Substitutions
\end{minipage} & \begin{minipage}[b]{\linewidth}\raggedright
Deletions
\end{minipage} & \begin{minipage}[b]{\linewidth}\raggedright
Insertions
\end{minipage} & \begin{minipage}[b]{\linewidth}\raggedright
S\% of ref words
\end{minipage} & \begin{minipage}[b]{\linewidth}\raggedright
D\%
\end{minipage} & \begin{minipage}[b]{\linewidth}\raggedright
I\%
\end{minipage} \\
\midrule\noalign{}
\endhead
\bottomrule\noalign{}
\endlastfoot
exp\_001 (base, CV20) & 20.30\% & 123 / 867 & 24 / 867 & 29 / 867 &
14.2\% & 2.8\% & 3.3\% \\
exp\_009 (small, CV24) & 24.42\% & 148 / 778 & 20 / 778 & 22 / 778 &
19.0\% & 2.6\% & 2.8\% \\
\end{longtable}
}

Substitutions account for 70\% of word errors in exp\_001 and 78\% in
exp\_009. Deletions and insertions are comparable across both models and
both test sets. The substitution-heavy profile is consistent with
Pashto's morphological richness: many error cases involve words where a
single phoneme difference (retroflex vs.~plain, or a vowel quality
distinction not written in the script) produces a completely different
lexical item rather than a partial character match.

Utterance-level accuracy is 53\% for exp\_001 (53 of 100 utterances with
zero errors after normalisation) and 34\% for exp\_009 (34 of 100). The
lower utterance accuracy for exp\_009 despite similar WER per word is
partly a dataset effect: CV24 utterances in error cases average 8.4
reference words versus 9.0 for CV20, meaning shorter utterances are more
sensitive to a single substitution.

\begin{center}\rule{0.5\linewidth}{0.5pt}\end{center}

\subsection{C.2 Error categories}\label{c.2-error-categories}

\textbf{Word-final morphological variants.} The most frequent category
in both models is one-word substitutions at phrase or sentence
boundaries where the correct word differs from the hypothesis in a
single character representing a grammatical suffix. Examples from
exp\_001:

\begin{itemize}
\tightlist
\item
  REF: \texttt{لاړجن\ لېونی} / HYP: \texttt{لاړجن\ لېونه} --- final
  \texttt{ی} (masculine indefinite) replaced by \texttt{ه} (feminine or
  alternative stem form); WER 50\%
\item
  REF: \texttt{دا\ مي\ مور\ ده} / HYP: \texttt{دا\ مې\ مور\ ته} ---
  predicate copula \texttt{ده} replaced by locative marker \texttt{ته};
  WER 50\%
\end{itemize}

These errors reflect substitutions between near-homophones that are
distinguished by short vowels or final consonants. Pashto does not write
short vowels, so the distinction must be inferred from acoustic context
and language model probability. Both models confuse these pairs at
similar rates.

\textbf{Retroflex and phoneme-inventory errors.} Pashto's retroflex
series (ټ, ډ, ړ, ڼ) is acoustically close to the corresponding
non-retroflex consonants (ت, د, ر, ن). Several substitution errors
involve this confusion. From exp\_009:

\begin{itemize}
\tightlist
\item
  REF: \texttt{زه\ څلوېښت\ ورځې\ وروسته\ خپل\ کلي\ ته\ ځم} / HYP:
  \texttt{زه\ سلوېښت\ وزې\ وروسته\ خپل\ کلي\ ته\ ځم} --- \texttt{څ}
  (Pashto-specific retroflex affricate) transcribed as \texttt{س}; WER
  38\%
\item
  REF: \texttt{کوچی\ راغلی\ دی} / HYP: \texttt{کوچۍ\ راغلی\ دی} ---
  masculine \texttt{ی} vs feminine \texttt{ۍ} suffix; WER 33\%
\end{itemize}

The first example is a Pashto-specific phoneme (\texttt{څ}, /ts/) that
has no counterpart in Arabic or Persian and for which whisper-base has
limited pre-training exposure.

\textbf{Function word deletion.} Deletions (2.6--2.8\% of reference
words) are dominated by Pashto function words: the genitive marker
\texttt{د}, the future particle \texttt{به}, the complementiser
\texttt{چې}, and the additive particle \texttt{هم}. These are short,
unstressed, and phonologically reduced in natural speech, making them
prone to deletion under acoustic uncertainty. Example from exp\_009:

\begin{itemize}
\tightlist
\item
  REF: \texttt{د\ بامیان\ په\ ښار\ کې} / HYP:
  \texttt{دهمیانو\ په\ ښار\ کې} --- \texttt{د} merged with the following
  noun and the noun itself distorted; effectively a deletion plus
  substitution pair
\end{itemize}

\textbf{Loanword and code-switched segments.} Utterances containing
English loanwords or Latin-script tokens produce the highest error rates
in both models. From exp\_001:

\begin{itemize}
\tightlist
\item
  REF: \texttt{د\ لپټاپ\ ونډوز\ می\ چلان\ activate\ کړه} / HYP:
  \texttt{داوړ} --- the model collapsed a nine-word sentence containing
  \texttt{activate} to a single unrelated word; WER 100\%
\end{itemize}

The loanword density in CommonVoice Pashto is higher in CV24 than CV20
(reflecting increased community contribution of tech-domain sentences),
which may contribute to the higher substitution rate in exp\_009.

\textbf{Acoustic hallucination.} A small proportion of utterances (4 of
100 for exp\_001, 3 of 100 for exp\_009) exhibit WER above 100\%,
meaning the model produces more output tokens than the reference
contains. These cases arise when the model loses confidence
mid-utterance and fills the remaining context with high-probability
Pashto phrases unrelated to the audio. The worst exp\_001 case:

\begin{itemize}
\tightlist
\item
  REF: \texttt{حضرت\ عثمان\ غنی} / HYP:
  \texttt{ښه\ د\ ده\ يا\ مسلمان\ نه\ يې\ دى} --- 3 reference words, 8
  hypothesis words; WER 267\%
\end{itemize}

These hallucinations account for disproportionate WER contribution
despite their low frequency.

\begin{center}\rule{0.5\linewidth}{0.5pt}\end{center}

\subsection{C.3 exp\_001 vs exp\_009
comparison}\label{c.3-exp_001-vs-exp_009-comparison}

The substitution rate increases from 14.2\% to 19.0\% moving from
exp\_001 (CV20 base) to exp\_009 (CV24 small). This appears
counterintuitive given that exp\_009 has a larger model (307M vs 74.4M
parameters) and more training data (113h vs 60h). Two factors explain
the discrepancy. First, the CV24 test set has a longer tail of dialectal
and domain-varied utterances than CV20; the substitution rate measures
how often the model produces the wrong word, not how wrong the word is,
and CV24's greater dialect range creates more ambiguous decision
boundaries. Second, exp\_009 has fewer insertions (2.8\% vs 3.3\%),
suggesting the larger model is more conservative about generating extra
content when uncertain --- trading some insertions for substitutions
compared to the base model. The net effect is that exp\_009's WER is
slightly higher on its test set (24.42\%) than exp\_001's on its own
(20.30\%), consistent with the full test-set results (24.89\% vs
21.22\%).

Both models show the same rank ordering of error categories:
substitutions \textgreater{} insertions \textgreater{} deletions, with
the same dominant failure patterns (morphological suffix confusion,
retroflex phoneme errors, function word deletion). This consistency
suggests that the error distribution is driven primarily by Pashto's
linguistic properties rather than by model size or training data volume.

\protect\phantomsection\label{refs}

\end{document}